%% file: main.tex
\documentclass{article}
\usepackage{ijcai26}

\usepackage{times}
\usepackage{soul}
\usepackage{url}
\usepackage[hidelinks]{hyperref}
\usepackage[utf8]{inputenc}
\usepackage[small]{caption}
\usepackage{graphicx}
\usepackage{amsmath}
\usepackage{amsthm}
\usepackage{amssymb}
\usepackage{amsfonts}
\usepackage{mathtools}
\usepackage{booktabs}
\usepackage{algorithm}
\usepackage{algorithmic}
\usepackage[switch]{lineno}
\usepackage{booktabs,makecell,array}
\usepackage{xcolor}
\usepackage{nicefrac}
\usepackage{siunitx}

\title{LLM-Guided Monte Carlo Tree Search over Knowledge Graphs: Composing Mechanistic Explanations for Drug--Disease Pairs}

\author{
Rishabh Jakhar$^1$
\and
Michel Dumontier$^1$\and
Remzi Celebi$^{1}$ 
\affiliations
$^1$Institute of Data Science, Department of Advanced Computing Sciences, Maastricht University, The Netherlands\\
\emails
rishabh.jakhar@maastrichtuniversity.nl
}

\begin{document}

\maketitle

\begin{abstract}
    Extracting multi-step explanations from knowledge graphs poses a combinatorial challenge requiring both heuristic guidance (as candidates proliferate with depth) and credit assignment (as path quality emerges over extended sequences). Frontier LLMs, strong on knowledge/reasoning benchmarks, offer a compelling source of such heuristics, yet their knowledge comes sans guarantees and compositional performance degrades as chains lengthen. We thus present \textsc{Tessera}, a 3-part neuro-symbolic framework that uses LLMs in a circumscribed role: for local discriminative judgement rather than autonomous multi-step generation; the knowledge graph then defines the hypothesis space enforcing hard structural constraints, and MCTS coordinates the long-horizon search with principled credit assignment via backpropagation. LLMs perform dual roles as a prior policy biasing exploration and a comparative state evaluator supplying reward signals. Evaluation on drug mechanism elucidation across two complementary knowledge graphs demonstrates fidelity to curated biology while surfacing coherent alternative mechanisms, with ablations confirming discriminative contribution from both LLM components. Beyond its current application, our framework offers a general paradigm for compositional reasoning over structured knowledge.\footnote{Code and supplementary material available at \url{https://github.com/RishabhJakhar/tessera}.}
\end{abstract}

\input{sections/intro}
\input{sections/rel_work}
\input{sections/methods}

\input{sections/exp}

\input{sections/results}
\input{sections/conclusion}

\section*{Acknowledgments}
This work was supported by the GENIUS Lab
(Generative Enhanced Next-Generation Intelligent Understanding Systems),
a collaboration between Delft University of Technology,
Maastricht University, DSM-Firmenich, and KickstartAI,
and by the NWO Long-Term Programme ROBUST,
initiated by the Innovation Center for Artificial Intelligence (ICAI).
We thank Dennis Soemers for helpful discussions on Monte Carlo Tree Search.

\bibliographystyle{named}
\bibliography{references}

\end{document}

%% file: sections/intro.tex
\section{Introduction}
Computational scientific discovery has engaged researchers for over half a century ~\cite{simon_scientific_1966}, often viewed as a heuristic search problem akin to planning or game-playing, but with the attendant expectation that its outputs constitute not merely predictions but explanations, expressed in formalisms amenable to inspection, challenge, and communication ~\cite{langley_integrated_2024}. Systems medicine exemplifies this: in drug repurposing, for instance, a putative statistical association between a drug and a disease is far more compelling when accompanied by a mechanistic account of how the intervention propagates, through biological intermediates, to produce its therapeutic effect.

Knowledge graphs, as explicit symbolic representations of domain knowledge, offer a natural substrate for such explanations ~\cite{Contextscontradictionsroadmapcomputational_sosa_2022a,daquin_role_2025}. Extracting these, however, poses a combinatorial challenge as candidates proliferate with depth, precluding exhaustive enumeration. The problem is thus well framed as heuristic search, with background knowledge guiding exploration towards promising candidates ~\cite{langley_integrated_2024}.

Frontier large language models, strong on biomedical knowledge and reasoning benchmarks ~\cite{phan_humanitys_2025,wang_mmlu-pro_2024,rein_gpqa_2023}, present a compelling source of such guidance. Yet LLMs exhibit a kind of \textit{approximate omniscience}: broad coverage of domain knowledge, but without guarantees of correctness ~\cite{kambhampati2023polanyi,kambhampati_position_2024}, subject to hallucinations and confabulations ~\cite{huang_survey_2025}. Empirically, compositional performance degrades systematically as task complexity increases: models succeed at single / few-step judgments but fail to compose them reliably as errors compound ~\cite{dziri_faith_2023}. Together, these recommend a circumscribed role: LLMs as sources of local discriminative judgment, not autonomous generators of extended reasoning chains. 

Assembling these local judgments into a multi-step explanation requires credit assignment: determining which early choices contribute to eventual explanation quality. Monte Carlo Tree Search provides a principled framework for this, accumulating evidence across trajectories and backpropagating credit over extended sequences. This yields a three-pillar framework, where the knowledge graph defines the hypothesis space and enforces hard structural constraints, the LLM provides soft, local semantic judgment, and MCTS coordinates long-horizon search. We refer to our framework as \textsc{Tessera} (\textbf{T}ree-search for \textbf{E}xplanation \textbf{S}ynthesis via \textbf{S}emantic \textbf{E}valuation and \textbf{R}anked \textbf{A}ctions).

Our contributions include: \textbf{(i)} a neuro-symbolic framework coupling MCTS with LLM-derived heuristics for extracting mechanistic explanations from biomedical knowledge graphs; \textbf{(ii)} a listwise prior policy (for exploration bias) that scales to large action sets through batched evaluation, exploiting LLMs' strength at relative judgment while avoiding long-context degradation; \textbf{(iii)} a comparative state evaluator (for reward signal) that scores partial paths against a depth-bound competitor set, using token-level probabilities for uncertainty-aware estimates and conditioning on accepted paths to assess marginal contribution; and \textbf{(iv)} evaluation on two complementary substrates: DrugMechDB (allowing deterministic evaluation against curated ground-truth), and the Multi-scale Interactome (enabling high-branching search, but without ground-truth, motivating an LLM-as-judge protocol, which we control and validate for reliability).

%% file: sections/rel_work.tex
\section{Related Work}
We situate our work relative to two strands of prior research: \textbf{(i)} \textit{Post-hoc attribution}, with methods such as Kelpie~\cite{rossi_kelpie_2022}, adversarial explanations~\cite{betz_adversarial_2022}, and GNNExplainer~\cite{ying_gnnexplainer_2019}, produces saliency-style explanations that highlight influential inputs (triples, subgraphs, or feature subsets), but often lack key explanation desiderata (e.g., coherence, parsimony); an inadequacy emphasised in broader critiques of attribution~\cite{chou_counterfactuals_2022} and carried over to link-prediction explainability~\cite{nunes_rew_exp_2025}. \textbf{(ii)} \textit{Path-based reasoning} explains predictions through multi-hop traversals that serve as interpretable traces. The dominant paradigm casts traversal as RL-based sequential decision-making: MINERVA~\cite{das_go_2018} rewards terminal arrival alone, with subsequent work progressively shaping rewards towards explanatory virtues with metapath compliance (PoLo;~\cite{liu_neural_2021}), phenotype associations (CoCo;~\cite{stork_explainable_2023}), and degree-based specificity heuristics (REx;~\cite{nunes_rew_exp_2025}). REx goes furthest, explicitly invoking explanatory qualities—fidelity, relevance, simplicity—yet still relies on structural proxies optimised through fixed-length rollouts (typically 3-4 hops). Outside RL, methods like DrugCORpath~\cite{song2025llmintegrated} incorporate LLM-derived embeddings of precomputed paths with subsequent clustering and classification, but remain constrained to certain metapaths at fixed, short lengths (2-3 hops). Our approach differs by replacing the hand-crafted proxies with a direct semantic evaluation via LLMs, and embedding this signal within a search algorithm (rather than policy rollouts), enabling variable depth over longer horizons. We view these as complementary: proxy rewards and policy rollouts offer efficiency; semantic evaluation and combinatorial search offer richer assessment and longer horizons, but at attendant cost. 

%% file: sections/methods.tex
\section{Methodology}
\subsection{Problem Setting}
Given a graph $\mathcal{G}$ as substrate, we formulate the problem as one of extracting an explanatory subgraph $g(d,z)$ that accounts for the therapeutic effect of a drug $d$ on a target disease $z$. We perform a lookahead search on $\mathcal{G}$: starting from $d$ and conditioned on $z$, the search traverses nodes and relations to uncover plausible mechanistic pathways.

The substrate is a directed, typed, multi-relational knowledge graph 
$
\mathcal{G} {=} (\mathcal{V}, \mathcal{R}, \mathcal{E}),
$
where $\mathcal{V}$ are entity nodes, $\mathcal{R}$ relation types, and $\mathcal{E} {\subseteq} \mathcal{V} {\times} \mathcal{R} {\times} \mathcal{V}$ the set of typed edges.

A search state $s {\in} \mathcal{S}$ is encoded as
$
s {=} (u, H, z),
$
where $u {\in} \mathcal{V}$ is the current node in $\mathcal{G}$, $H {\in} \mathcal{E}^{\ast}$ is the ordered path traversed to reach $u$, and $z {\in} \mathcal{V}$ is the target node. $V(H)$ denotes the nodes appearing along $H.$ For any given search, $z$ remains fixed. Since states are path-dependent, visiting the same graph node via different paths corresponds to distinct search states. The legal actions at state $s$, $\mathcal{A}(s)$, form a subset of the outgoing edges of $u$ and are determined in two stages. The first is a filtering on the top-$k$ neighbours ranked by a personalised PageRank score $\mathrm{ppr}_z(v)$, with a one-hot personalisation vector encoding the target disease $z$:
\[
\mathcal{A}_k(s) = \operatorname{Top}\text{-}k_{\text{\,ppr}_z} \{ (r,v) : (u,r,v) \in \mathcal{E}, v \notin V(H)\}.
\]
The second is an augmentation to $k,$ ensuring adequate representation of key node types: for a designated set $\mathcal{T} {\subseteq} \mathcal{V}$ and target proportion $\lambda {\in} (0,1]$, if nodes in $\mathcal{T}$ comprise less than proportion $\lambda$ of the targets in $\mathcal{A}_k(s),$ additional actions with $v {\in} \mathcal{T}$ are injected from the ranked tail, up to an augmentation budget $\tau$. Writing $\mathcal{A}^+_{\tau}(s)$ for the resulting set of injected actions, the final action set is
\[
\mathcal{A}(s) = \mathcal{A}_k(s) \cup \mathcal{A}^+_{\tau}(s), \quad k \le |\mathcal{A}(s)| \le k{+}\tau.
\]
Since an action represents the tuple $(r,v),$ the transition function is deterministic: executing $(r,v)$ in $s{=}(u,H,z)$ yields afterstate $s' {=} (v, H\! +\!\!\!\!\!+  (u,r,v),\; z).$

\subsection{Search Algorithm}
We use Monte Carlo Tree Search (MCTS), guided by an LLM-based prior policy $\pi_{\text{LLM}}({\cdot} {\mid} s)$ to bias exploration towards high-probability actions and a comparative, path-aware state evaluation $v(s_L)$ to assign leaf values.

The search tree consists of states $s$ as nodes, with outgoing edges $(s{,}a)$ for all $a {\in} \mathcal{A}(s)$. Each edge stores statistics: visit count $N(s{,}a)$, cumulative action value $W(s{,}a)$, mean action value $Q(s{,}a)$, and a prior probability $P(s,a) {\coloneqq} \pi_{\text{LLM}}(a{\mid} s)$. Search starts at root $s_0{=}(d{,}\varnothing{,}z)$, and evolves under a preset simulation budget $T$. Each simulation cycles through 4 phases:

\paragraph{Selection.} Starting at the root $s_0$, the tree is traversed until a leaf $s_L$ is reached. At each step $t{<}L,$ an action $a_t$ is chosen to maximise $Q(s_t,a) {+} U(s_t,a)$ using PUCT \cite{rosin_multi-armed_2011,silver_mastering_2017}, with 
\[
U(s_t,a) = c_{\text{puct}}\,
P(s_t,a)\,\frac{\sqrt{\sum_b N(s_t,b)}}{1+N(s_t,a)}.
\]
We formulate the exploration coefficient $c_{\text{puct}}{=}\delta(d)\cdot\phi(N)$ to be depth and visit-count dependent, where
\[
\delta(d) = \frac{c_0}{1+\alpha d}\quad\text{and}\quad \phi(N)=1+\beta e^{-N/K}.
\]
Here $d$ is the depth and $N$ the total visit-count of the parent state $s_t.$ $\delta(d)$ decays and hence tempers exploration as depth increases, while $\phi(N)$ provides an early-visit exploration bonus that vanishes as $s_t$ gets more visits.

\paragraph{Evaluation.}
On reaching $s_L{=}(u,H,z)$, we check for terminal states: if $u {=} z$ (target state), the corresponding path history is admitted to the explanation set $\mathcal{H}_{\text{exp}}(s_0),$ incoming edge $(s_{L-1},a_{L-1})$ is closed and $v(s_L)$ is set to 0. If $\mathcal{A}(s_L){=}\varnothing$, i.e. a dead-end, again edge $(s_{L-1},a_{L-1})$ is closed with $v(s_L){=}0.$ For non-terminal states (excluding root $s_0)$, a scalar value $v(s_L){\in}[-1,1]$ is obtained from the LLM-based state evaluator (section \ref{methods:state_eval}). 

\paragraph{Backpropagation.}
A backward pass is made for steps $t {\leq} L$ with updates: $ N(s_t,a_t) {\leftarrow}  N(s_t,a_t){+}1,$ $W(s_t,a_t) {\leftarrow} W(s_t,a_t){+}v(s_L),$ and $Q(s_t,a_t) {\leftarrow} \frac{W(s_t,a_t)}{N(s_t,a_t)}.$

\paragraph{Expansion.}
Leaf $s_L$ is expanded unconditionally. Expansion is atomic over (i) materialising all successors in $\mathcal{A}(s_L),$ and (ii) assigning priors $P(s_L{,}a){=}\pi_{\text{LLM}}(a {\mid} s_L).$ 

Upon search termination, subgraph $g(d,z)$ is constructed as the edge union of all paths in $\mathcal{H}_{\text{exp}}(s_0)$.

\subsection{Prior Policy}
\label{methods:prior_policy}
We estimate $\pi_{\text{LLM}}({\cdot} {\mid} s)$  by pairing an LLM with a listwise sorting procedure. The design draws upon ranking algorithms in information retrieval and in particular, adapts the multi-pivot, $n$-ary quicksort of \cite{godfrey_likert_2025} for action scoring on graphs. 

Four considerations shape the method: \textbf{(i)} as $|{\mathcal{A}(s)}|$ typically exceeds the quality cap for a single LLM pass, we use batched evaluations, with a batch size $W {\ll} |\mathcal{A}(s)|,$ to avoid long-context degradation (e.g. position bias~\cite{liu_lost_2024}), \textbf{(ii)} considering LLMs are stronger at relative than absolute judgements~\cite{sun_is_2023,qin_large_2024}, we perform $n$-ary listwise comparisons, which batching naturally enables by presenting sets of actions jointly, \textbf{(iii)} to maximise throughput given LLM latency, batches are scored concurrently, with quicksort requiring only a light-weight final aggregation to scale batch-level judgements to a global distribution over $\mathcal{A}(s),$ and \textbf{(iv)} since MCTS is most sensitive to ordering of the top actions, the procedure progressively truncates the action set, concentrating computation on the distribution head. The full procedure is formalised in Algorithm 1 (Suppl. S3). We outline the key steps below:

\paragraph{Batch construction.} The quicksort procedure runs over $m$ passes, truncating the working action set $A_T$ over a linear schedule from $T_1{=}|\mathcal{A}(s)|$ to $T_m{=}W$.  In each pass, we select $k{=}\lfloor W/2\rfloor$ quantile-spaced pivots from $A_T$ using bootstrap scores $\boldsymbol{\beta}_T$\footnote{for $m{=}1,$ $\boldsymbol{\beta}_T$ are the PageRank scores $\mathrm{ppr}_z(\cdot),$ for $m {\geq} 2$ these are replaced with the batch-standardised LLM scores $\tilde s(a).$};  batches are then formed by adjoining this shared pivot set $P$ (size $k)$ to disjoint chunks of non-pivots (size ${\leq} W{-}k)$.
\paragraph{LLM evaluation.} Each batch $B{\in} \mathcal{B}$ is sent concurrently for LLM evaluation, with context including the parent leaf state $s{=}(u,H,z)$, rubric $\mathcal{R}_{\text{prior}}$, and for every action $a{=}(r,v){\in} B$, the predicate $r$ and node $v$ attributes (\textit{identifier}, \textit{type}, \textit{label}, and \textit{description}). The model returns a rank $\pi_B(a)$ and score $s_B(a)$ per action, and on the final pass (i.e. for the top-$W$ actions) a free-text justification to support interpretability\footnote{Full prior policy prompt (template and example) in Suppl. S2.}. A within-batch z-score standardisation yields $\tilde s_B(a)$.
\paragraph{Cross-batch aggregation \& global order.} Within each ordered batch, we identify the left and/or right pivot for every non-pivot. To obtain a transitive global order, we use the shared pivots as anchors: each pivot’s rank and score are averaged across batches to form an anchor score $S(p){=}(-\bar\pi(p),\bar s(p))$. We then extend $S(\cdot)$ to non-pivots by interpolating between their neighbouring pivots. Having established $S(a)$ for every action, the final ordering is obtained by sorting lexicographically on $(S(a),\, s_B(a),\, \beta(a))$.
\paragraph{Prior over actions.} Given the final ordering, each action is assigned a utility that blends its normalised rank with its batch-standardised LLM score. We then map these utilities to probabilities via a temperature-controlled softmax. 

\subsection{State Evaluation}
\label{methods:state_eval}
We perform a comparative evaluation of non-terminal leaves: each candidate is (i) assessed jointly with a small set of competing paths $\mathcal{C}(s_L)$, and (ii) scored for its marginal contribution relative to the current explanation set $\mathcal{H}_{\text{exp}}.$

Formally, consider the evaluation of a leaf state $s_L{=}(u{,}H{,}z)$ under root $s_0.$ Let $\mathcal{L}(s_0)$ be the set of all active search states under $s_0,$ and $H_s$ the root-to-$s$ path, whose length $|H_s|$ gives the depth of state $s.$ 

We aim to estimate the state value $v(s_L){\in}[-1,1]$ using an LLM-based evaluator by scoring its path history $H_{s_L}$ against a \emph{depth-bound} competitor set $\mathcal{C}(s_L).$ Competitors are selected from a filtered pool $\mathcal{F}_{\Delta}(s_L),$  containing only states within a depth window $\Delta\in\mathbb{N}_0$ of $s_L:$ 
\[
\mathcal{F}_{\Delta}(s_L)\;=\;\Big\{\,s\in \mathcal{L}(s_0)\setminus\{s_L\}\ :\ \big|\,|H_s|-|H_{s_L}|\,\big|\le \Delta \,\Big\}.
\]
$\mathcal{C}(s_L)$ is then constructed as $\{s_L\} \,  \cup \, \operatorname{Top}\text{-}k_{\prec} (\mathcal{F}_{\Delta}(s_L))$, where the ordering $\prec$ is lexicographic (descending)  on two search-tree statistics:
\[
\begin{aligned}
\log P_{\mathrm{path}}(s) & =\sum_{(s,a)\in H_s}\log\!\big(\max\{P(s,a),\varepsilon\}\big), \quad \text{and} \\
N_{\mathrm{cum}}(s) &= \sum_{(s,a)\in H_s} N(s,a).
\end{aligned}
\]

Here $N$ and $P$ are per-edge visit counts and priors, and $\varepsilon{>}0$ is a small numerical clamp. If fewer than \(k\) competitors exist, all are included; if none exist, \(\mathcal{C}(s_L)=\{s_L\}\).

Each state \(s{\in}\mathcal{C}(s_L)\) is presented to the model through its path history $H$, with the prompt conditioned on the current explanation set $\mathcal{H}_{\text{exp}}(s_0){=}\{H {:} \exists \, s{=}(z,H,z)\}.$ The model evaluates each state using a predefined ordinal rubric $\mathcal{R_{\text{state-eval}}}{=}\{\rho_{\min}{<}{\dots}{<}\rho_{\max}\}{\subset}\mathbb{Z}$ of integer labels.\footnote{Full state-eval prompt (template and example) in Suppl. S4.} Instead of using the output label directly, we obtain a continuous, uncertainty-aware score by taking a softmax over the rubric, using the model’s token-level log-probabilities $\ell(\rho_i {\mid} s)$, to get a categorical probability distribution  $p(\rho_i {\mid} s)$. The resulting expected score,  $\sum_i p(\rho_i {\mid} s)\,\rho_i,$ is normalised to get $v(s):$
\[
v(s) = \frac{2\,\sum_i p(\rho_i \mid s)\,\rho_i - (\rho_{\max}+\rho_{\min})}{\rho_{\max}-\rho_{\min}} \in [-1,1]
\]

\paragraph{Non-LLM baseline (PPR-eval).}
As a deterministic baseline, we replace the LLM evaluator with a value computed from the target-conditioned PageRank scores $\mathrm{ppr}_z(\cdot)$. Since $\mathrm{ppr}_z$ is not on the $[-1,1]$ scale used in PUCT backups, we apply a global rank calibration:
\[
v_{\text{ppr}}(s{=}(u,H,z)) \;=\; 2\cdot \operatorname{rank\_pct}\!\big(\mathrm{ppr}_z(u)\big) {-} 1\in[-1,1],
\]
where $\operatorname{rank\_pct}{\in}[0,1]$ is the normalised rank of $\mathrm{ppr}_z(u)$ over all nodes.

%% file: sections/exp.tex
\section{Experimental Setup}
\subsection{Substrate Graphs}

\paragraph{\textbf{DrugMechDB (DMDB).}} A merged supergraph of expert-curated per-indication mechanisms from DrugMechDB ~\cite{gonzalez-cavazos_drugmechdb_2023}, allowing evaluation against ground-truth mechanisms, thus decoupling algorithmic performance from substrate completeness. The resulting graph (5,128 nodes; 10,064 edges) spans multiple levels of biological organisation (molecular activities, pathways, anatomical structures, phenotypes) but offers only a small per-step action set (out-degree mean=2; p90=4; max=224) and minimal strong connectivity (largest strongly connected component (SCC) covering just 1\% of nodes).

\paragraph{\textbf{Multi-scale Interactome (MSI).}}To complement the above and probe performance under high-branching search, we evaluate on a large-scale biological knowledge graph adapted from the Multiscale Interactome ~\cite{ruiz_identification_2021} (adaptation details in Suppl. S5). The graph (29,698 nodes; 921,953 edges) encompasses 17,527 proteins, 9,798 biological functions, 1,550 drugs, and 821 diseases. Its dense protein--protein interaction core and GO function hierarchy yield a much larger action space (out degree mean=31, p90=75, max=2,299). Moreover, the largest SCC spans 92\% of nodes, indicating extensive bidirectional reachability and, as search states are path-dependent, a high path multiplicity, making search more challenging.

\subsection{Language Models and Evaluation Set}

For state evaluation during search, we experiment with SOTA \textit{non-reasoning} models, \textsc{Gpt-4.1}, \textsc{DeepSeek-V3.1}, and \textsc{Qwen3-235B}, on the premise that decomposition of search into local discriminative judgments at least partially obviates the need for reasoning models. The prior policy uses \textsc{Gpt-4.1} throughout; priors are cached per search state, enabling reuse across runs varying only the state-eval model. Model identifiers, inference settings, and algorithm hyperparameters are provided in Supplement S1.

Due to the computational costs associated with LLM inference (see Supplement S10 for a per-search LLM-call bound), we limit our evaluation to a representative sample of 15 drug-disease pairs. To ensure diversity across pharmacological and pathophysiological contexts, pairs were sampled across 12 MeSH disease categories and 9 ATC therapeutic classes, and further stratified by DMDB reference graph size (node/edge count). Full details in Supplement S6.

\subsection{Evaluation Protocol}
\subsubsection{With DrugMechDB as substrate}
The predicted subgraphs are compared deterministically against their curated (gold) counterparts along 4 axes (formal definitions in Suppl. S7).

\paragraph{Node Set Agreement (NSA).} Set overlap between the curated and predicted nodes; order agnostic.

\paragraph{Edge Set Agreement (ESA@$h$).} Edge agreement with $h$-hop tolerance. A predicted edge $(u{\to}v)$ counts for \textit{precision} if the curated graph connects $(u{\rightsquigarrow}v)$ within $\le h$ hops (i.e. shortcuts allowed). Conversely, a curated edge is \textit{covered} (counts for \textit{recall}) if the prediction connects $(u{\rightsquigarrow}v)$ within $\le h$ hops (i.e. detours allowed). ESA@1 reduces to strict edge overlap. 

\paragraph{Transitive-Closure Agreement (TCA).} Path independent reachability among curated nodes. Let $C_\star$ and $C_p$ be the transitive closures of the curated and predicted graphs over the curated node set $V_\star$ (i.e., $C_\star, C_p \subseteq V_\star \times V_\star$). Precision is the fraction of pairs in $C_p$ that also lie in $C_\star$; recall is the fraction of pairs in $C_\star$ that also lie in $C_p$ (predicted paths may traverse any, including non-curated, nodes). Captures agreement of the overall connectivity structure.

\paragraph{Exact Path Agreement (EPA).} Overlap of \emph{exact} drug to disease paths, evaluated only at path lengths present in the curated subgraph. \textbf{EPA--IV} \textit{(in-vocabulary)} scores only those predicted paths that stay entirely within curated nodes, making precision more lenient to off-mechanism nodes and isolating \emph{verbatim fidelity} to the curated narrative. \textbf{EPA--OW} \textit{(open-world)} scores all predicted paths, yielding stricter precision and emphasizing \emph{parsimony} by penalizing paths that route through non-curated nodes.

\begin{figure*}[t]
    \centering
    \includegraphics[width=0.95\linewidth]{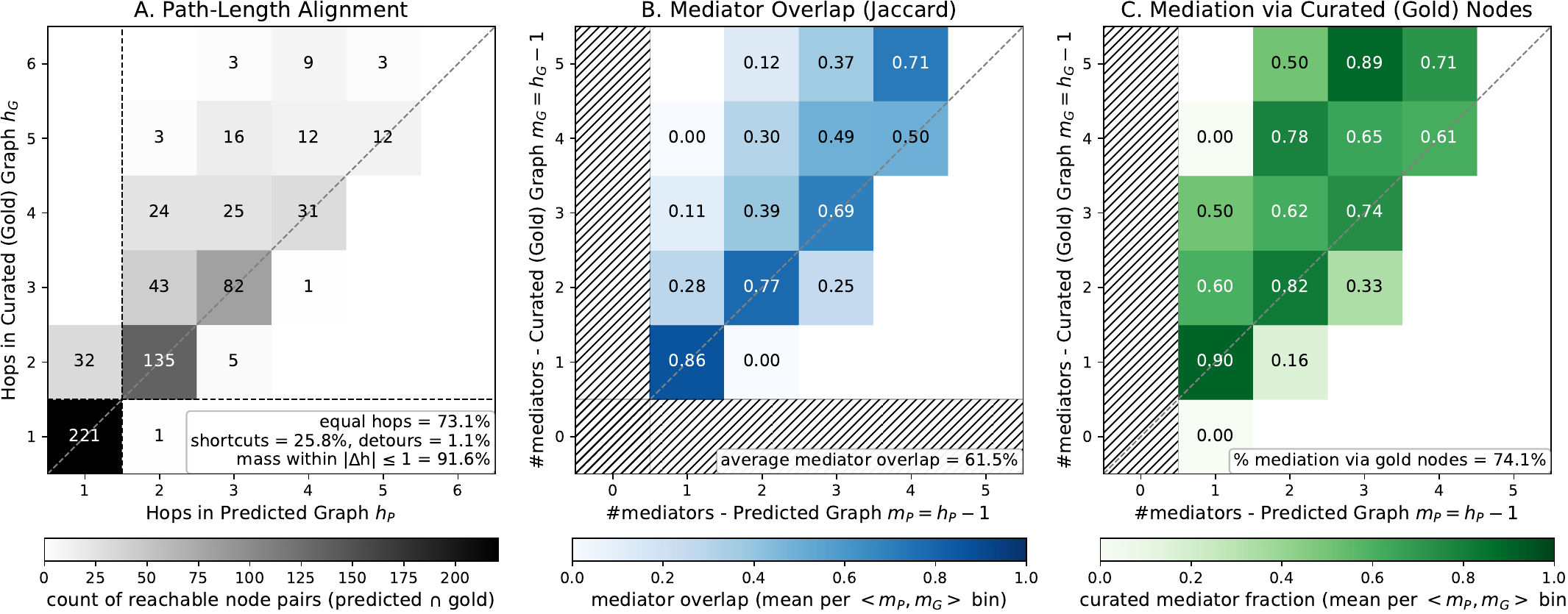}
    \caption{\textbf{Panel A}: joint distribution of shortest-path hop counts ${(h_P{,}h_G)}$ between node pairs reachable in both curated and predicted graphs; diagonal = matched length, above = shortcuts, below = detours; $h{=}1$ separates direct vs. mediated connections. \textbf{Panel B}: Jaccard overlap of predicted vs. curated mediator sets (mean per $m_P{,}m_G$ bin); undefined for $m_P{=}0$ or $m_G{=}0$ (hatched). \textbf{Panel C}: fraction of mediators in the \emph{predicted} paths that are \textit{curated (gold)} nodes (mean per $m_P{,}m_G$ bin); undefined for $m_P{=}0$ (hatched).}
    \label{fig:3_panel_mediator}
\end{figure*}

\subsubsection{With the Multi-scale Interactome as substrate}
In absence of ground-truth mechanisms to compare against, we implement a rubric-based LLM-as-judge protocol, using large language models augmented with expert-curated reference knowledge to score the predicted subgraphs. Each subgraph is judged on 5 dimensions--\emph{Biological Plausibility}, \emph{Mechanistic Coherence}, \emph{Contextual Specificity}, \emph{Completeness}, \emph{Conciseness}--with each dimension rated on a 1-5 ordinal scale. As reference, we use the mechanism-of-action entries from DrugBank~\cite{DrugBank60DrugBankKnowledgebase_knox_2024} and mechanism graphs from DrugMechDB to ground assessments (prompt and rubric in Suppl. S8).

To control for scoring variability, we employ a $3{\times}3$ design: \textbf{(i)} as LLMs can exhibit sensitivity to the presentation-order of graph elements, each judge evaluates 3 different serialisations (permutations) of the same graph, varying only in node/edge ordering; \textbf{(ii)} as different LLMs can exhibit different scoring behaviours, each subgraph is independently scored by 3 judges (\textsc{Gpt-4.1}, \textsc{DeepSeek-V3.1}, \textsc{Qwen3-235B}). For each individual judgment, we extract the model’s token probabilities over the five ordinal scale values $k{\in}\{1,{\dots},5\}$, renormalise them, and compute a probability-weighted expected rating $\mathbb{E}[s]{=}\sum_{k} k\,p({k})$, yielding a continuous score in $[1,5]$. The reported scores \textit{(per subgraph, dimension)} are the mean of these expected ratings across all 9 permutation–judge combinations.

%% file: sections/results.tex
\section{Results and Discussion}

\subsection{Evaluation on DrugMechDB}

\subsubsection{Aggregate performance} Table~\ref{tab:dmdb-results} reports scores pooled across state-eval LLMs. Node-level agreement is strong (NSA Micro P/R = $0.71/0.83$). At edge level, ESA@1 (no hop slack) yields Micro P/R = $0.44/0.64.$ Allowing for a 1-hop tolerance (ESA@2) raises precision ($0.44$ to $0.50$), while recall remains flat, indicating more 1-hop shortcuts than detours in the predicted subgraphs (examined further below). To investigate whether the modest precision at both $h{=}\{1,2\}$ reflects outright spurious edges or plausible but uncurated mediators, a valid possibility given DMDB mechanisms are not exhaustive, we evaluated ESA while \textit{restricting edges to curated nodes only}, which raised Micro P to $0.86$ at $h{=}1$ and $0.99$ at $h{=}2,$ confirming the low rate of false positives (spurious edges) among curated nodes. Comparing reachability between curated node pairs in both graphs, TCA shows Micro P/R = $0.99/0.66,$ i.e. nearly all reachability claims made by the algorithm are valid, with the predictions missing about $34\%$ of node pairs reachable in gold (primarily due to missing nodes). Finally, comparing complete drug${\to}$disease paths, EPA-IV has a Micro P of $0.84,$ implying the algorithm's traversal of curated nodes by and large follows gold paths. On the other hand, the significant drop in precision from EPA-IV to EPA-OW ($0.84$ to $0.27$) underscores that predicted paths frequently incorporate nodes outside the curated mechanism.

\newcommand{\microcell}[2]{\makecell[r]{#1\\\footnotesize(#2)}}      
\newcommand{\macrocell}[3]{\makecell[r]{#1\\\footnotesize(#2)\\\footnotesize[\,#3\,]}} 

\begin{table}[tb]
  \centering
  \setlength{\tabcolsep}{3pt}
  \renewcommand{\arraystretch}{1}
  \begin{tabular*}{\columnwidth}{@{\extracolsep{\fill}}lrrrr@{}}
    \toprule
    \textbf{Axis} & \makecell{\textbf{Micro P}} & \makecell{\textbf{Micro R}} & \makecell{\textbf{Micro F1}} & \makecell{\textbf{Macro F1}} \\
    \midrule
    \textbf{NSA}     & \microcell{0.71}{0.68--0.74} & \microcell{0.83}{0.77--0.88} & \microcell{0.76}{0.73--0.80} & \microcell{0.77}{0.73--0.80} \\
    \textbf{ESA@1}   & \microcell{0.44}{0.39--0.49} & \microcell{0.64}{0.54--0.74} & \microcell{0.52}{0.46--0.58} & \microcell{0.53}{0.47--0.60} \\
    \textbf{ESA@2}   & \microcell{0.50}{0.46--0.56} & \microcell{0.64}{0.55--0.73} & \microcell{0.56}{0.51--0.62} & \microcell{0.57}{0.51--0.63} \\
    \textbf{TCA}     & \microcell{0.99}{0.98--1.00} & \microcell{0.66}{0.58--0.75} & \microcell{0.79}{0.73--0.85} & \microcell{0.81}{0.75--0.87} \\
    \textbf{EPA--IV} & \microcell{0.84}{0.72--0.95} & \microcell{0.33}{0.21--0.48} & \microcell{0.48}{0.33--0.63} & \microcell{0.44}{0.30--0.58} \\
    \textbf{EPA--OW} & \microcell{0.27}{0.18--0.37} & \microcell{0.33}{0.21--0.48} & \microcell{0.30}{0.20--0.41} & \microcell{0.31}{0.21--0.42} \\
    \bottomrule
  \end{tabular*}
  \caption{DrugMechDB results (15 pairs ${\times}$ 3 state-eval LLMs = 45 runs). Columns show Micro (pooled across runs) and Macro (per-run mean) aggregates. Cells display point estimate on line~1; 95\% CI on line~2. Confidence intervals via run-level bootstrap.}
  \label{tab:dmdb-results}
\end{table}

\subsubsection{Path-length and mediator analysis} 
Figure \ref{fig:3_panel_mediator} decomposes the structural agreement. \textbf{A.} (Path-Length Alignment) shows \(73.1\%\) of mass on the diagonal (matched hops) and \(91.6\%\) within \(|\Delta h|{\le} 1\). When not matched, paths tend to compress (above diagonal (shortcut) = \(25.8\%\)) rather than expand (detour = \(1.1\%\)), reinforcing why ESA precision improves with \(h{=}2\) while recall stays flat; \textbf{B.} (Mediator Overlap) shows that for pairs requiring at least one mediator ($h{\geq}2$ hops), predicted paths share on average \(61.5\%\) of their mediators with the corresponding gold paths (Jaccard overlap); the remainder includes substitutions involving both curated and uncurated nodes; \textbf{C.} (Mediation via Gold Nodes) shows that \(74.1\%\) of all predicted mediators are curated nodes, thus most mediation mass flows through curated biology even if exact routes differ.

The comparisons above show the algorithm largely identifies and connects the correct endpoints (high NSA/TCA), often via slightly shorter routes (Panel A) that nonetheless pass through curated biology (Panel C). It also shows good fidelity to curated paths (high EPA-IV precision) but can lack parsimony, proposing paths that traverse uncurated mediators (low EPA-OW precision). While understandable, given the search uses pre-trained LLMs with no task-specific supervision to match DMDB exactly, it does raise the question of whether these uncurated proposals constitute valid alternatives.

\subsubsection{Qualitative analysis}
We examined 15 subgraphs (\textsc{Gpt-4.1} state-eval) and identified recurring divergence patterns that nonetheless preserved biological coherence; per-pair annotations in Suppl. S9. Two dominant patterns appeared: (1) \textit{mechanistic elaboration}, where predictions added upstream signaling pathways or regulatory detail (12/15 pairs; e.g., IL-1-mediated signaling pathway, adenylate cyclase-GPCR signaling, smooth muscle contraction), and (2) \textit{phenotype insertion}, where predictions introduced explicit phenotypic states between biological processes and disease endpoints (9/15 pairs; e.g., Hypertension before Hypertensive disorder, joint swelling before Rheumatoid arthritis). Less frequent patterns included insertion/omission of chemical mediators (5/15; e.g., Dopamine, Aldosterone, Calcium) and addition/omission of anatomical/cellular context (4/15; e.g., lymphoid system, Leukocyte). In all cases, direction of effect and endpoint alignment were maintained. Figure \ref{fig:benazepril} illustrates this on a representative pair, \textit{Benazepril}-\textit{Hypertensive Disorder}, which otherwise shows poor performance on exact path metrics (EPA-(IV \& OW)=0). Both graphs share the canonical ACE inhibitor pharmacology (Benazepril ${\to}$ Benazeprilat ${\to}$ inhibits ACE ${\to}$ \mbox{$\downarrow\!$ Angiotensin II).} The curated graph then routes through \textit{renal sodium absorption} and \textit{water retention}; the prediction instead captures an upstream regulator in the same axis, \textit{aldosterone secretion}. The prediction also inserts an explicit phenotypic intermediate (Hypertension). 

\definecolor{myblue}{HTML}{DAE8FC}
\definecolor{mygreen}{HTML}{D5E8D4}
\definecolor{mygrey}{HTML}{E0E0E0}
\begin{figure}[tb]
    \centering
    \includegraphics[width=.95\linewidth]{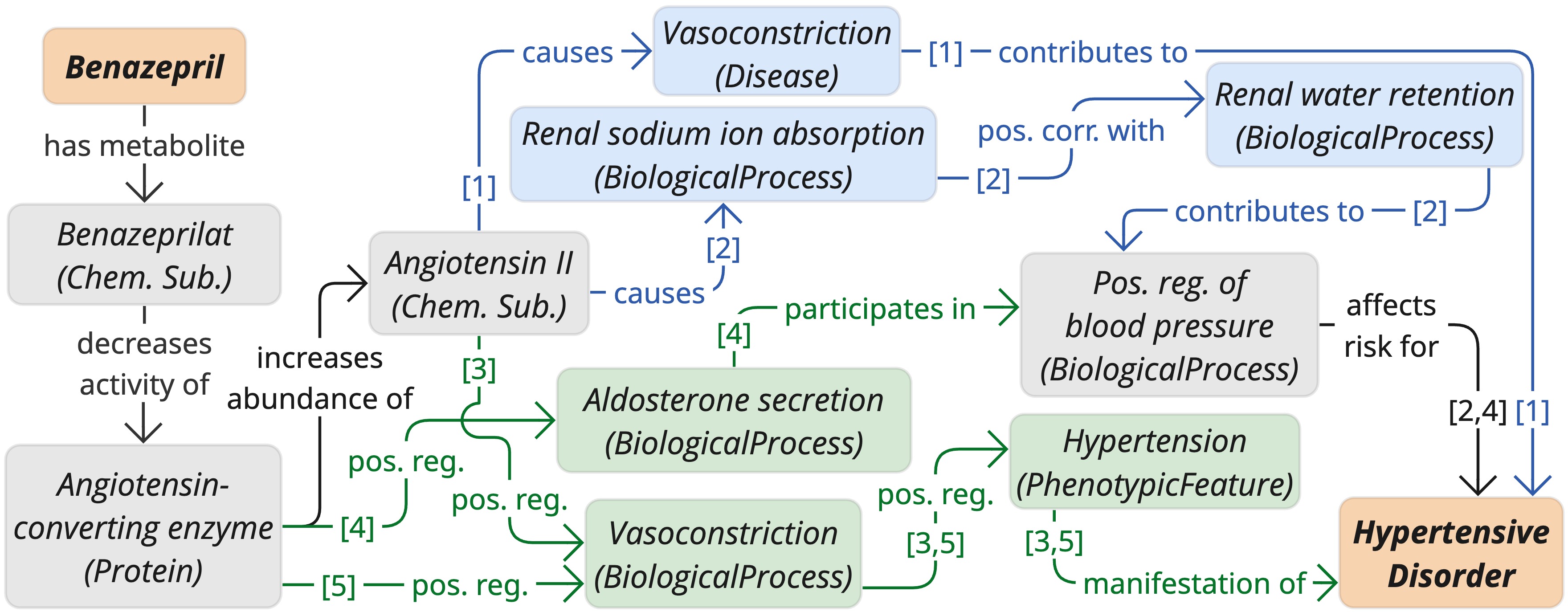}
    \caption{Predicted vs. Curated explanatory subgraph for $\mathsf{Benazepril{\to}Hypertensive\_Disorder}$. Legend: \fcolorbox{black}{mygrey}{\rule{0pt}{1.2ex}\rule{1.2ex}{0pt}} Common; \fcolorbox{black}{myblue}{\rule{0pt}{1.2ex}\rule{1.2ex}{0pt}} Curated only; \fcolorbox{black}{mygreen}{\rule{0pt}{1.2ex}\rule{1.2ex}{0pt}} Predicted only.}
    \label{fig:benazepril}
\end{figure}

\subsection{Evaluation on the Multi-scale Interactome}
Following the LLM-as-judge protocol, let $s^{m}_{g,d}$ be the mean expected score across the 3x3 serialisation-judge combinations, for subgraph $g,$ dimension $d,$ and state evaluation model $m\,{\in}\,$$\{\text{\textsc{Gpt-4.1}},$ $\text{\textsc{DeepSeek-V3.1}},$ $\text{\textsc{Qwen3-235B}}\}.$ 

\paragraph{Reliability.} To assess the reliability of the scoring process, we estimated intra- and inter-rater agreement using two-way mixed-effects, absolute-agreement, intraclass correlations (ICC) ~\cite{koo_guideline_2016}. Within-model agreement across the 3 serialisations was extremely high, ICC(3,3) = 0.989 (\textsc{DeepSeek-V3.1}), 0.987 (\textsc{Qwen3-235B}), and 0.981 (\textsc{Gpt-4.1}), indicating practical invariance to permutation order. Likewise, agreement among the 3 judge LLMs, pooled across the 5 dimensions, was excellent: ICC(3,3)=0.946, 95\% CI 0.93-0.96, justifying the use of $s^{m}_{g,d}$ as a reliable input. 

\begin{figure}[tb]
    \centering
    \includegraphics[width=.95\linewidth]{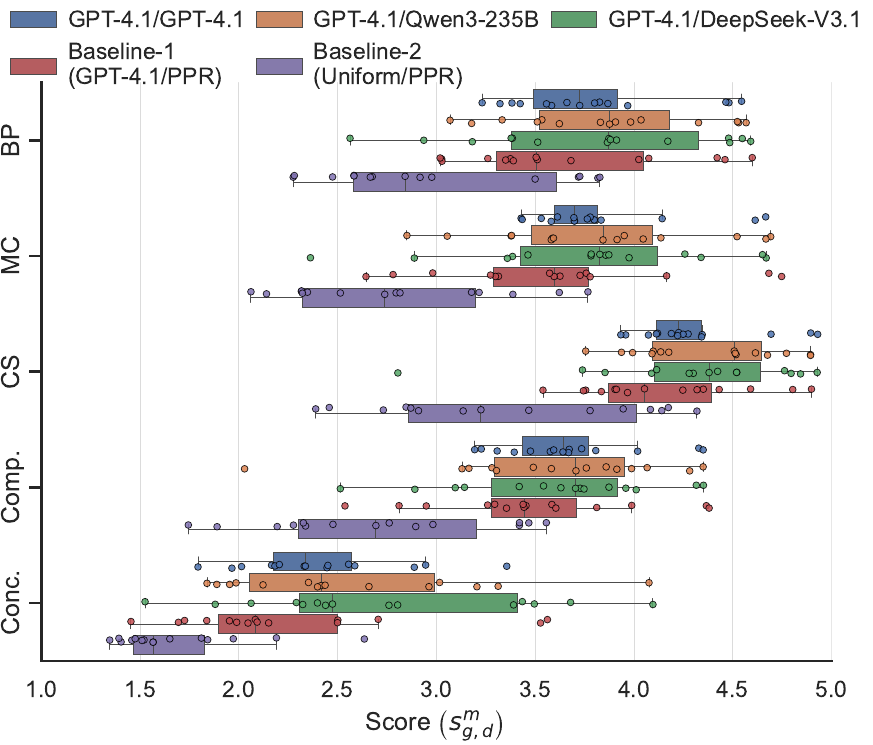}
    \caption{Score ($s^{m}_{g,d}$) distributions on MSI across 5 dimensions (BP: Biological Plausibility; MC: Mechanistic Coherence; CS: Contextual Specificity; Comp.: Completeness; Conc.: Conciseness). Legend shows prior model/state-eval model.}
    \label{fig:msi_dist}
\end{figure}

\paragraph{Score distributions.}Figure~\ref{fig:msi_dist} shows the score distributions; included are two baselines that substitute the LLM-based state evaluation with PPR-eval. Baseline-1 retains the LLM prior; Baseline-2 pairs PPR-eval with a uniform prior over legal actions. Across the three LLM-based state-eval variants, \textit{Contextual Specificity} was uniformly highest (medians spanning 4.23-4.51), followed by \textit{Biological Plausibility} (3.72-3.87), \textit{Mechanistic Coherence} (3.70-3.84), and \textit{Completeness} (3.64-3.71); \textit{Conciseness} was scored the lowest (2.34-2.47). \textsc{Gpt-4.1} exhibited the tightest dispersion (IQR between 0.22-0.42); \textsc{Qwen3-235B} and \textsc{DeepSeek-V3.1} showed wider spread (IQR between 0.54-1.10). While both baselines scored lower across all dimensions, Baseline-2 collapsed markedly, i.e. the exploration bias induced by the LLM-informed priors largely offsets PPR-eval; absent this, search tends to degenerate and disperse across the target-proximal space. 

\paragraph{Cross-model comparison.} We compute the difference between dimension-averaged scores, per drug-disease pair,  for each state-eval model pair $
\overline{\Delta}_{g}({m_1,m_2}) = \frac{1}{5}\sum_d (s^{m_1}_{g,d} - s^{m_2}_{g,d}),$ and report (Table \ref{tab:msi_pair_agreement}) the median (signed and absolute) and fraction of subgraphs scored within $\delta{=}\nicefrac{1}{4}$ and $\nicefrac{1}{2}$ of a scale point. Additionally, we measure rank concordance across subgraphs using Kendall’s $\tau_b.$ Looking at the results, signed medians are near-zero; absolute medians span 0.13–0.16, small relative to the 1-5 rubric scale. Agreement within-$\delta$ further underscores the tight clustering, with \textsc{DeepSeek-V3.1} accounting for most outliers. Rank concordance is also strong ($\tau_{b}{=}0.60{-}0.79$), implying the models largely agree on which drug-disease pairs are easier or harder to explain, in part due to topological constraints of the shared substrate. 

\begin{table}[b]
  \centering
  \renewcommand{\arraystretch}{1}
  \begin{tabular}{lrrrr}
    \toprule
    Model Pair &
    \makecell[c]{Median \\ $\overline{\Delta}_{g}\,/\,|\overline{\Delta}_{g}|$} &
    \makecell[c]{$\tau_b$} &
    \makecell[c]{$\Pr(|\overline{\Delta}_{g}|{\le} \delta)$ \\ $\delta\!: 0.25 / 0.50$} \\
    \midrule
    
    \makecell[l]{\textsc{Gpt-4.1} vs \\ \textsc{DeepSeek-V3.1}} & $-0.10\,/\,0.13$ & $0.79$ & $0.67\,/\,0.80$ \\
    
    \makecell[l]{\textsc{Qwen3-235B} vs \\ \textsc{DeepSeek-V3.1}} & $0.02\,/\,0.16$ & $0.60$ & $0.67\,/\,0.87$ \\

    \makecell[l]{\textsc{Gpt-4.1} vs \\ \textsc{Qwen3-235B}} & $-0.02\,/\,0.15$ & $0.77$ & $0.87\,/\,1.00$ \\
    \bottomrule
  \end{tabular}
  \caption{Cross-model (state-eval) agreement on dimension-averaged subgraph scores.}
  \label{tab:msi_pair_agreement}
\end{table}

\begin{figure}[tb]
    \centering
    \includegraphics[width=.975\linewidth]{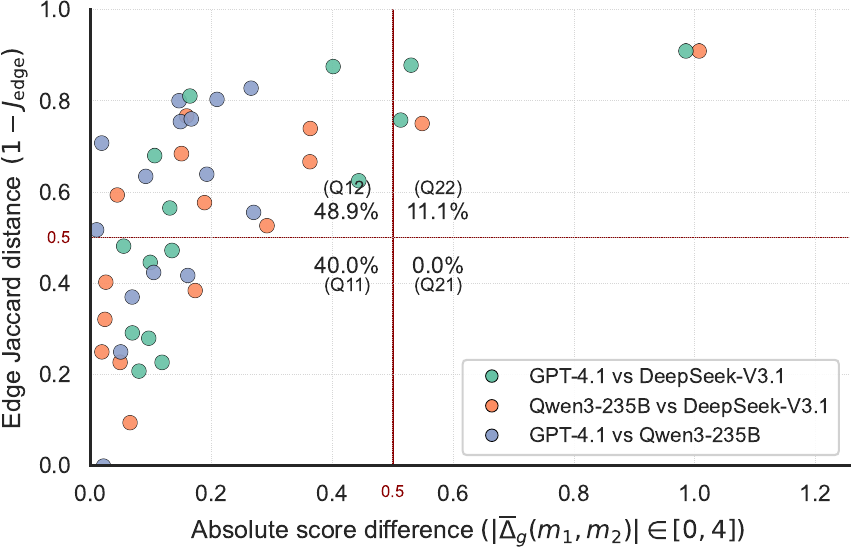}
    \caption{Per-subgraph absolute score difference vs. structural divergence across state-eval model pairs; quadrants show pooled percentages. }
    \label{fig:msi_struc_vs_score}
\end{figure}

\paragraph{Structural convergence vs. Mechanistic pluralism.} The patterns above imply strong cross-model agreement in the scored outcomes. For insight into the extent to which this reflects structural convergence versus distinct but equally defensible mechanisms, Figure~\ref{fig:msi_struc_vs_score} relates the absolute score difference ($|\overline{\Delta}_{g}|$) to structural divergence (edge-level Jaccard distance). We see a substantial structural spread (edge-distance IQRs between $0.37{-}0.41$ across model pairs), pointing to significant mechanistic pluralism between the models, and, in turn, to the discriminative influence of the state-eval function on the search. Taking a per-quadrant view, using $|\overline{\Delta}_{g}| {=} 0.5$ (half a scale step) and edge-distance ${=} 0.5$ as thresholds, mass concentrates comparably between $\text{Q}11$ ($40.0\%$, \textit{convergent mechanisms}) and $\text{Q}12$ ($49.0\%$, \textit{plural mechanisms}). Notably, Q21 (structural convergence, score divergence) is empty across pairs, consistent with the high ICC values and a reliable subgraph evaluation methodology.

\subsection{Ablation Study: Prior Policy}
\begin{figure}[tb]
    \centering
    \includegraphics[width=.975\linewidth]{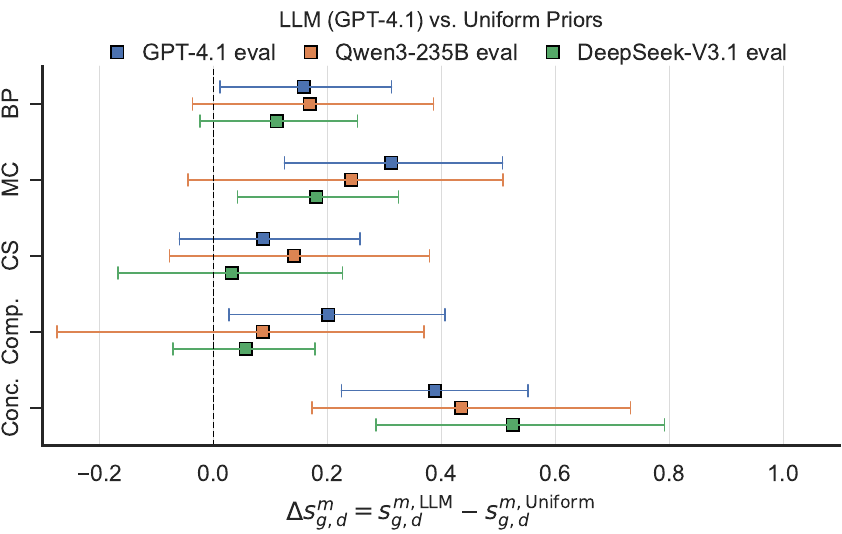}
    \caption{Prior policy ablation: mean score difference (LLM minus uniform prior) by dimension and state-eval model; 95\% bootstrap CIs.}
    \label{fig:ablation_msi}
\end{figure}

To assess the influence of the prior policy, we conduct an ablation on MSI, replacing $\pi_{\text{LLM}}(\cdot{\mid}s)$ utilising \textsc{Gpt-4.1} with a uniform distribution over legal actions. Figure~\ref{fig:ablation_msi} shows, for each dimension $d$ and state-eval model $m$, the mean of score differences $\Delta s^{m}_{g,d} {=} s^{m,\text{LLM}}_{g,d} {-} s^{m,\text{Uniform}}_{g,d}$ across drug–disease pairs.  Overall, the LLM-derived priors yield higher scores across rubric dimensions and state-eval models. The largest gains are on \emph{Conciseness}, followed by \emph{Mechanistic Coherence}, and smallest on \emph{Contextual Specificity}, the latter indicative of both priors, bound by the same PPR-filtered neighbourhood, exploring similar regions of the substrate; larger differences in conciseness and coherence thus appear to reflect how paths are assembled within this overlapping region. 

To examine this, we compute four structural metrics per subgraph: drug${\to}$disease path count ($n_{\text{path}}$), mean path length ($\ell_{\text{path}}$), fraction of paths consisting solely of protein--protein interactions ($f_{\text{PPI-only}}$), and ratio of biological-process to protein nodes ($r_{\text{BP:Prot}}$). Table~\ref{tab:path_structure} reports averages across drug-disease pairs and state-eval models. To situate the analysis in the substrate's topology, consider that MSI is protein-dominated: protein–protein interactions comprise 84\% of all edges, and proteins outnumber biological-process nodes 17{,}527 to 9{,}798, predisposing any under-discriminative policy to proliferate protein-heavy subgraphs. The search under a uniform prior manifests this bias clearly, admitting on average $2.5{\times}$ more paths per subgraph, that are substantially more protein-dominated ($\bar{f}_{\text{PPI-only}}$: $0.39$ vs. $0.13$) and markedly less process-enriched ($\bar{r}_{\text{BP:Prot}}$: $0.29$ vs. $0.64$). In contrast, $\pi_{\text{LLM}}$ biases search onto a smaller set of longer, process-mediated routes, suppressing many shorter, redundant, protein-only chains. Taken together with the score deltas in Figure~\ref{fig:ablation_msi}, this partially explains why explanations become simultaneously more complete and more concise, and why mechanistic coherence improves despite the increase in average path length.

\newcommand{\meansd}[2]{#1\,{\scriptstyle \pm #2}}

\begin{table}[tb]
  \centering
  \renewcommand{\arraystretch}{1}
  \setlength{\tabcolsep}{2.5pt}
  \begin{tabular}{lrrrr}
  \toprule
  \makecell[c]{Prior} &
  \makecell[c]{$\bar{n}_{\text{path}}$}&
  \makecell[c]{$\bar{\ell}_{\text{path}}$} &
  \makecell[c]{$\bar{f}_{\text{PPI-only}}$} &
  \makecell[c]{$\bar{r}_{\text{BP:Prot}}$} \\
  \midrule
  \textsc{Gpt-4.1} & $\meansd{7.58}{5.28}$ & $\meansd{7.89}{2.67}$ & $\meansd{0.13}{0.17}$ & $\meansd{0.64}{0.24}$ \\
  \textsc{Uniform} & $\meansd{19.56}{9.68}$ & $\meansd{5.97}{0.78}$ & $\meansd{0.39}{0.30}$ & $\meansd{0.29}{0.19}$ \\
  \bottomrule
\end{tabular}
  \caption{Subgraph structural metrics: LLM vs. Uniform priors. Entries show mean ${\pm}$ SD across drug-disease pairs, state-eval models.}
  \label{tab:path_structure}
\end{table}

%% file: sections/conclusion.tex
\section{Conclusion}
Extracting multi-step explanations from knowledge graphs requires heuristic guidance to manage combinatorial explosion and credit assignment to evaluate extended sequences. To address these, we present \textsc{Tessera}, a neuro-symbolic decomposition wherein the knowledge graph bounds the hypothesis space, LLMs supply local semantic judgment, and MCTS coordinates search with principled backpropagation. Evaluation on drug mechanism elucidation demonstrates the approach recovers established biology while surfacing coherent alternatives. The decomposition applies broadly to compositional reasoning over structured knowledge.

\textit{A note on limitations.} Our evaluation spanned 15 drug-disease pairs; while stratified for diversity, broader validation is needed. Additionally, a systematic hyperparameter optimisation remains non-trivial and unexplored given LLM inference costs and sparse ground-truth signal; current settings reflect manual tuning from pilot experiments.